\documentclass[10pt,twocolumn,letterpaper]{article}

\usepackage{cvpr}
\usepackage{times}
\usepackage{epsfig}
\usepackage{graphicx}
\usepackage{amsmath}
\usepackage{amssymb}
\usepackage{multicol}

\usepackage[pagebackref=true,breaklinks=true,letterpaper=true,colorlinks,bookmarks=false]{hyperref}

\cvprfinalcopy 

\def\cvprPaperID{****} 

\begin{document}

\title{Positional Attention for Efficient BERT-Based Named Entity Recognition}

\author{Mo Sun, Siheng Xiong, Yuankai Cai, Bowen Zuo \\
Georgia Institute of Technology\\
\texttt{\{msun330, sxiong45, ycai321, bzuo6\}@gatech.edu}
}

\maketitle

\begin{abstract}
This paper presents a framework for Named Entity Recognition (NER) leveraging the Bidirectional Encoder Representations from Transformers (BERT) model in natural language processing (NLP). NER is a fundamental task in NLP with broad applicability across downstream applications. While BERT has established itself as a state-of-the-art model for entity recognition, fine-tuning it from scratch for each new application is computationally expensive and time-consuming. To address this, we propose a cost-efficient approach that integrates positional attention mechanisms into the entity recognition process and enables effective customization using pre-trained parameters. The framework is evaluated on a Kaggle dataset derived from the Groningen Meaning Bank corpus and achieves strong performance with fewer training epochs. This work contributes to the field by offering a practical solution for reducing the training cost of BERT-based NER systems while maintaining high accuracy.
\end{abstract}

\section{Introduction}

This project uses a language representation model called Bidirectional Encoder Representations from Transformers (BERT) to solve a common problem in natural language processing (NLP). The BERT model can be used in NLP for classification, predicting whether a sentence is grammatically correct or not, and many other applications. Unlike recent language sensation models \cite{brown1992class}, BERT pre-trains deep representations from unlabeled text using the context of the text around it \cite{devlin2018bert}. The pre-trained BERT model can then be fine-tuned to suit specific applications by using another dense layer.

The task of named entity recognition (NER) is to recognize the named entities in the given text. For example, it can classify words as names, verbs, etc. NER is a preliminary and important task in NLP and can be used in many downstream NLP tasks, such as relation extractions \cite{bunescu2005shortest}, event extraction \cite{chen2015event}, and question answering \cite{yao2014information}. In recent years, numerous methods have been studied for NER tasks, including Hidden Markov Models (HMMs) \cite{bikel1997nymble}, Support Vector Machines (SVMs) \cite{isozaki2002efficient}, and Conditional Random Fields (CRFs) \cite{lafferty2001conditional}. 

The BERT model is a state-of-the-art model in NLP to do entity recognition in a Google project, and the training set of this model is the Wikipedia dataset. So, when training the model, Google used huge storage for the whole dataset and Tpu clusters to handle this task in a few days. However, if we use it in an application, we need to customize this model to our dataset. It is easy to collapse and hard to get ideal results with the pre-trained parameters. We decided to build a framework that doesn’t need to train the BERT model from scratch (using pre-trained parameters) and can get good results by using fewer training epochs. One challenge of doing this is that the BERT model is large and expensive to train and fine-tune, so one must be careful to choose methods that have low enough time and space costs. 

This project is important because to apply natural language models in different situations, a lot of companies ask the Google cloud solution team to help them train a model to fit their needs. Although Google can train models based on customers' needs, it is also a high cost and slow processing to build the framework for each company. So, we want to find a solution to reduce the cost of training the process of the BERT model. The research gap we are filling is to introduce positional attention mechanisms in entity recognition and try to find a low-cost method to customize models with pre-trained parameters. Therefore, our project will demonstrate a way that the BERT model can easily be customized to a particular need or context efficiently. 

This project used a dataset on Kaggle which is extracted from the Groningen Meaning Bank corpus that is tagged, annotated, and built specifically to train the classifier to predict named entities such as name, location, etc. This dataset contains a huge amount of combinations of words and is updated regularly, which is helpful for us to train the model.

\section{Methodology}

In our study, we utilize the BERT model, which is based on the state-of-the-art methods described in the papers “Attention is All you Need” and “Show, Attend and Tell: Neural Image Caption Generation with Visual Attention.” The transformer architecture relies on attention mechanisms, eliminating the need for recurrence. Attention mechanisms help determine the focus or emphasis placed on each word in a sentence while considering each individual word. The pre-trained BERT model converts each word in a sentence into Word Piece embeddings and processes them through 12 layers of encoders, each containing a multi-headed self-attention layer and a feed-forward neural network, as shown in figure 1.

\begin{figure}[h]
  \centering
  \includegraphics[width=0.8\linewidth]{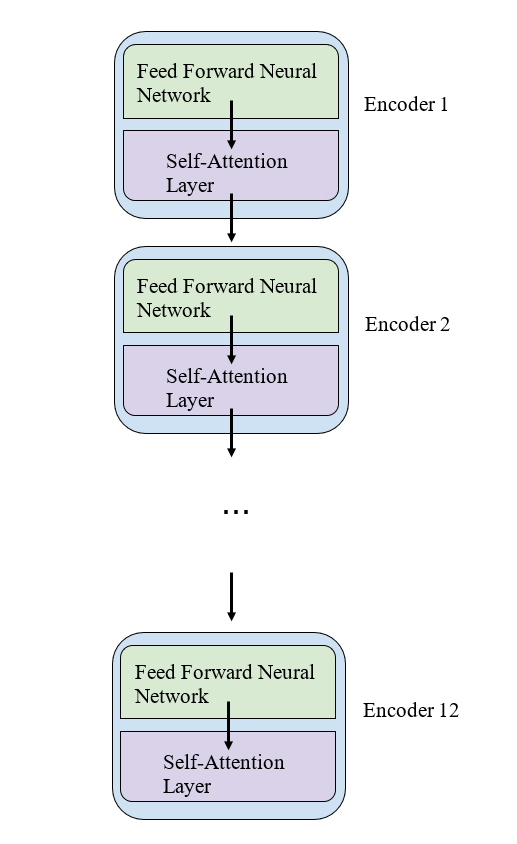}
  \caption{BERT’s 12 layers of two-part encoders}
  \label{fig:image_label}
\end{figure}

The Word Piece embeddings are how each word is represented in BERT \cite{devlin2018bert}. Each word is a vector of 768 features that describes the meaning of the word. Words with similar meanings will have embeddings that are close to each other, such as “couch” and “sofa,” whereas words with different meanings, such as “tree” and “spaceship” will have embeddings that are numerically far apart from each other. BERT accounts for the relative positions of word encodings, enabling it to differentiate between sentences like "Dogs chase cats" and "Cats chase dogs." BERT is pre-trained on various tasks, such as masked language modeling, where some words are replaced by a [MASK] token, and the model has to predict the original word.
In this project, we build upon the pre-trained BERT model to address Named Entity Recognition (NER) for a specific text dataset from Kaggle. We leverage the pre-trained BERT model's knowledge while fine-tuning it for our domain-specific task. Our approach involves the following steps:

a. Preprocessing the text dataset: Tokenizing the sentences, converting words into Word Piece embeddings, and encoding their positions.

b. Fine-tuning the pre-trained BERT model: Adapting the model to our specific NER task by training it on our domain-specific dataset, which allows the model to learn the relevant vocabulary and context.

c. Evaluating the model: Assessing the performance of the fine-tuned BERT model on our NER task using standard evaluation metrics, such as precision, recall, and F1-score. Comparing the performance of our approach with other existing methods to understand its effectiveness in the given context.

We believed that using the BERT model for solving the NER problem would be successful for several reasons:

    1. Proven success in NLP tasks: The BERT model has shown great performance in various NLP tasks, outperforming many previous models. Its pre-training on large text corpus and attention mechanism has proven effective in understanding context and relationships between words.

    2. Transfer learning: The pre-trained BERT model provides a solid foundation that can be fine-tuned for domain-specific tasks. 

    3. Adaptability to context: Fine-tuning BERT for our specific NER task allows it to learn the relevant domain-specific vocabulary and context, making it more suitable for identifying named entities in our dataset.

The creative side of our approach mostly comes from these two parts below.

    1. Customized fine-tuning: Although BERT has been used for NER tasks before, our approach fine-tunes the model specifically for our domain and dataset, potentially leading to better performance in our problem domain.

    2. Exploration of different techniques: Our approach could involve experimenting with different fine-tuning techniques, model architectures, or optimization methods, which could potentially improve the model's performance on the NER task.

\section{Experiments}

In the preprocessing step, we added a [CLS] tag to collect information about the entire sentence as a whole, including meaning information and positional information after we downloaded the data.Then we used a BERT meta dictionary which helped us to transfer the different words into a unique number, therefore we got a high dimensional vector as the model input.

Specifically, found that some words weren’t in any category, so we didn’t focus on those words in our model. We masked those words as a [PAD] tag, which told the model to only get positional information in these words.
Third, we tried to handle the compatibility of the BERT meta dictionary. Sometimes, if the dictionary did not contain the word, the transferring function would split the word into more than one tag. Although they had the same label, we had to combine the answers when the result was returned.

After the model optimization, the best model we selected was at training epoch 12 and Learning rate 0.00003 that gave us the minimum training loss and valid loss.We total collected 500 sentences online as the test data and each sentences have different meaning and sentence structures. To evaluate and obvious the final output we print the final two output that included sentence and tokenized\_sentence.For the tokenized\_sentence that included BERT-token meaning and Grammer-token meaning. The example has been provided in Table \ref{tab:BERT} and Table \ref{tab:Grammer}. For example a sentence: Alice will go to China this Saturday! Her father works in WHO. The sentence will be split into ['Alice', 'will', 'go', 'to', 'China', 'this', 'Saturday!', 'Her', 'father', 'works', 'in', 'WHO', '.'] The BERT-token meaning and Grammer-token meaning output is  ['B-art' 'B-per' 'O' 'O' 'O' 'B-geo' 'O' 'B-tim' 'O' 'O' 'O' 'O' 'O' 'B-org' 'O' 'B-art'] and ['\$' 'NNP' 'VBD' 'PRP' 'VBZ' 'NNP' '.' '\$'].We use the function of metrics.flat f1 score(), and the final F1 score to measure the success of our model is 0.8143. We also implement two other models to compare the result.

We used the BERT model as the main information extraction function, then add some dropouts on each neural network layer that can help us to train the model with more flexibility. Table \ref{tab:model} shows the modelling.

\begin{table}
\begin{center}
\begin{tabular}{|l|c|c|}
\hline
Idx & Layer & Optimizer \\
\hline\hline
0 & BERT uncased model & AdamW(with Warmup) \\
1 & Fully connected layer & Dropout \\
2 & Categorical loss & -- \\
2 & Positional loss & -- \\
3 & Active loss & -- \\
\hline
\end{tabular}
\end{center}
\caption{Modeling.}
\label{tab:model}
\end{table}

\begin{table}
\begin{center}
    \begin{tabular}{|l|c|}
      \hline
      Token & Meaning \\
      \hline\hline
      B - art & attribute \\
      B - geo & geo-location \\
      B - time & time \\
      B - org & Positional loss \\
      \hline
    \end{tabular}
    \caption{BERT-token meaning comparison table}
    \label{tab:BERT}
\end{center}
\end{table}
  
\begin{table}
\begin{center}
\begin{tabular}{|l|c|}
  \hline
  Token & Meaning \\
  \hline\hline
  NN* & noun \\
  VB* & verb \\
  DT & daytime \\
  \hline
\end{tabular}
\caption{Grammar-token meaning comparison table}
\label{tab:Grammer}
\end{center}
\end{table}

In our problem's structure, the first two indexes are the main part that learned parameters. We used the AdamW optimizer with default parameters and we varied the learning rate over the course of training. This corresponds to increasing the learning rate linearly for the first warmup steps training steps. We used the batch size to determine the warmup steps which is equal to 2,500. And, we used the dropout rate = 0.1 on the fully connected layers.  

\begin{figure}[h]
\begin{center}
   \includegraphics[width=0.8\linewidth]{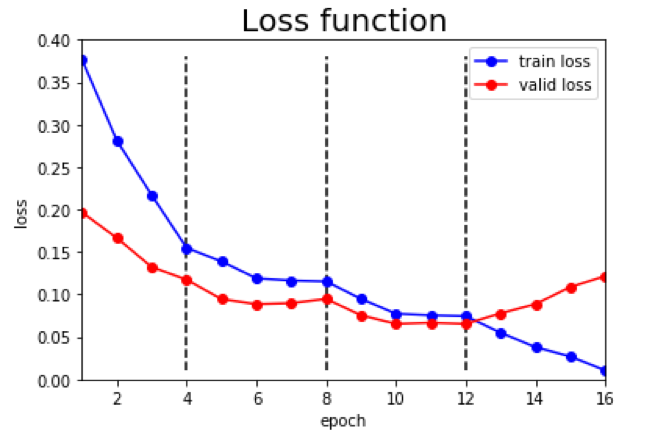}
\end{center}
   \caption{Loss function of the training process}
\label{fig: Loss_function}
\end{figure}

To view the performance of our model in the loss function, As shown in Fig. \ref{fig: Loss_function}, we can split the training process into 4 phases.

In phase 1, the training loss and validation loss dropped as a linear function, which means that our model tried to fit the data in the training warmup stage. In phase 2, the training loss stuck around 0.077, but the validation loss has a small fluctuation at the end of the phase 2, therefore we reduce the learning rate a little bit to avoid the loss plateau. In phase 3, the loss structure is similar to phase 2. In phase 4, we found that although we reduced the loss, the training loss decreased and the validation loss increased, which means the model was overfitted at this stage. Therefore, the best model we selected was at training epoch 12.

In this project, we presented the BERT model, the first sequence transduction model based entirely on attention, replacing the recurrent layers most commonly used in encoder-decoder architectures with multi-headed self-attention. 

Current techniques restrict the power of the pre-trained representations, especially for the fine-tuning approaches. The major limitation is that standard language models are unidirectional, and this limits the choice of architectures that can be used during pre-training. The BERT model alleviates the unidirectionality constraint by using a “masked language model” (MLM) pre-training objective. The masked language model randomly masks some of the tokens from the input, and the objective is to predict the original vocabulary ID of the masked word based only on its context. Unlike the left-to-right language model pre-training, the MLM objective enables the representation to fuse the left and the right context.

BERT has recently demonstrated to be quite good at sentence completion tasks if trained on a large corpus. A recently constructed sentence completion task, however, shows these models perform quite poorly when compared to humans if the sentence completion tasks require some common-world knowledge that cannot be gleaned from the corpus. Therefore, we require models that don't just learn word embeddings by being trained on input text, but also learn what those words mean in contexts, such as language learning and world modeling. This form of learning, called grounded language learning, is currently an active area of study.

We are excited about the future of attention-based models and plan to apply them to other tasks. We plan to extend the BERT to problems involving input and output modalities other than text and to investigate local, restricted attention mechanisms to efficiently handle large inputs and outputs such as images, audio, and video. Making generation less sequential is the research goal of ours.

\section{Discussion}

\textbf{Conditional Random Field (CRF)}: The input features are typically generated from the words in the sentence, such as their part-of-speech tags, morphological features, and contextual information. The labels are assigned to each word in the sentence based on the context of the word and its neighboring words, using a set of predefined rules and features.

The CRF model learns the conditional probability distribution of the output sequence given the input sequence using a maximum likelihood estimation algorithm. The model considers all possible label sequences and assigns a probability score to each one based on the input features and the conditional probabilities of the labels given the input features.

To be specific, features for one word include: the lowercase form of the word, the last three characters, the last two characters, a boolean indicating whether it is all uppercase, a boolean indicating whether it is a number, a boolean indicating whether it starts with a capital letter, and the part-of-speech tag.

If the word is not the first word in the sentence, the features also include: the lowercase form of the previous word, a boolean indicating whether the previous word is all uppercase, a boolean indicating whether the previous word is a number, a boolean indicating whether the previous word starts with a capital letter, and the part-of-speech tag of the previous word. In addition, the feature 'BEG' and 'END' are used to mark the beginning and end of the sentence, respectively.

$$
\begin{array}{cccccccc}
 \textbf {label} & \rightarrow &  \textbf {label} &  \textbf {weight} & \textbf {label} & \rightarrow &  \textbf {label} &  \textbf {weight}\\
 0 & \rightarrow &0 & 2.3892 & \text {org} & \rightarrow &\text {gpe} & -0.3517\\
 \text {eve} & \rightarrow &\text {eve} & 1.6219 & \text {art} & \rightarrow &\text {per} & -0.3586 \\
\text {art} & \rightarrow &\text {art} & 1.6178 & \text {art} & \rightarrow &\text {org} & -0.4206\\
\text {per} & \rightarrow &\text {per} & 1.3803 & \text {geo} & \rightarrow &\text {gpe} & -0.4370\\
 \text {tim} & \rightarrow &0 & 1.1758 & \text {gpe} & \rightarrow &\text {geo} & -0.4666\\
 0 & \rightarrow &\text {tim} & 1.0222 & \text { org } & \rightarrow &\text {per} & -0.5217\\
 \text {nat} & \rightarrow &\text {nat} & 0.9881 & \text {geo} & \rightarrow &\text {per} & -0.5921\\
 0 & \rightarrow &\text {nat} & 0.7800 & \text {per} & \rightarrow &\text {gpe} & -0.7292\\
 \text {nat} & \rightarrow &0 & 0.7692 & \text {geo} & \rightarrow &\text { org } & -1.1887\\
 \text {tim} & \rightarrow &\text {tim} & 0.7586 & \text {org} & \rightarrow &\text {geo} & -1.2461
\end{array}
$$

To get the features for a sentence, we first uses a pre-trained spaCy model to perform part-of-speech tagging on the input sentence and obtain the corresponding POS tags for each word in the sentence. Then, for each word in the sentence, we obtain a set of features for that word. We also split the string by whitespace to obtain a list of labels, where each label corresponds to one word in the sentence.

Finally, for the model, we use the function of sklearn\_cr-fsuite.CRF(). For the metrics, we use the function of metrics.flat\_f1\_score(), and the F1 score is 0.7783. We also show the top 10 likely and unlikely transitions.

\textbf{Transformer}: The model of Transformer is designed to process sequences of data, such as text, and has been used in a wide range of NLP tasks. The steps to use Transformer for NER include:
\begin{itemize}
    \item Data Preparation: Annotate the named entities in the text.

\item Preprocessing: Tokenize the text, encode the tokens using a pre-trained tokenizer, and convert the labels to numerical values.

\item Model Training: Fine-tune a pre-trained Transformer model on the NER task.

\item Evaluation: Evaluate the performance of the trained model on a held-out test set using metrics such as precision, recall, and F1-score.

\item Inference: Use the trained model to perform NER on new, unseen text data by tokenizing the text, encoding it using the pre-trained tokenizer, feeding it through the trained model, and decoding the predicted entity labels.
\end{itemize}

\begin{figure}[h]
\begin{center}
   \includegraphics[width=0.8\linewidth]{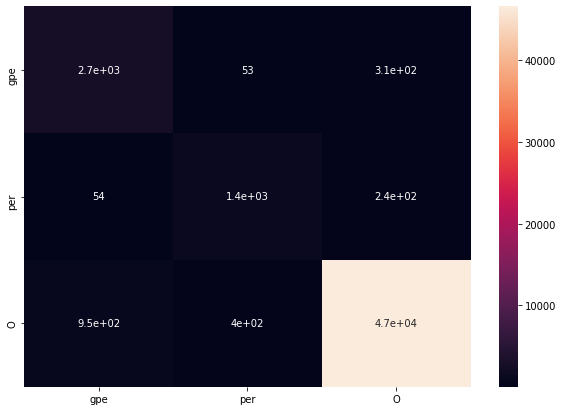}
\end{center}
   \caption{Confidence Matrix for Transformer}
\label{fig: Tran_conf_mat}
\end{figure}

To be specific, we first read Glove Embeddings (https://nlp.stanford.edu/projects/glove/). Then for data pre-processing, we map B-entity and I-entity to same entity, train only on PER and GPE, and compute token2idx, idx2token, tag2idx, idx2tag. We also pad the input using PAD\_IDX and pad the output with -1, after which all the sequences within a batch will have same length. The parameters of models are set as: embedding\_size = 300, num\_heads = 6, num\_layers = 6, dropout = 0.0, hidden\_dim = 2048. For training, the batch size is set as 64, and the number of epochs is 20, and we use optim.SGD(model.parameters(), lr=0.001, momentum=0.9) as optimizer. The split ratio for training and validation is 0.8:0.2. Our results show that the precision and recall for GPE are 0.7250 and 0.8806, the precision and recall for PER are 0.7501 and 0.8233, and the precision and recall for 0 are 0.9884 and 0.9717. Fig. \ref{fig: Tran_conf_mat} shows the confidence matrix of the model.

\textbf{Bidirectional Long Short-Term Memory (BiLSTM)}: Another popular deep learning model for NER is BiLSTM. The steps to use Transformer and BiLSTM for NER are similar. The only difference is model construction. To be specific, our BiLSTM model includes an Embedding layer, a Dropout layer, a Bidirectional LSTM layer and a TimeDistributed Dense layer. The embedding size is 104, the dropout rate is 0.1, the number of units in Bidirectional LSTM is 100, return\_sequence is True, and the recurrent\_dropout is 0.1, and the activation of TimeDistributed Dense is softmax. Our results show that the precision and recall for GPE are 0.6741 and 0.8256, the precision and recall for PER are 0.6923 and 0.7598, and the precision and recall for 0 are 0.9323 and 0.9418, which are slightly lower than those of Transformer.

\end{document}